\documentclass[10pt,a4paper]{article}

\usepackage{natbib}
\pdfoutput=1

\usepackage{multirow}

\usepackage{amsmath,amssymb}

\usepackage{authblk}

\title{Quantile Convolutional Neural Networks for Value at Risk Forecasting}
\author[1]{G\'abor Petneh\'azi\thanks{gabor.petnehazi@science.unideb.hu}}
\affil[1]{Doctoral School of Mathematical and Computational Sciences, University of Debrecen}
\date{}

\begin{document}
\maketitle

\begin{abstract}
This article presents a new method for forecasting Value at Risk. Convolutional neural networks can do time series forecasting, since they can learn local patterns in time. A simple modification enables them to forecast not the mean, but arbitrary quantiles of the distribution, and thus allows them to be applied to $VaR$-forecasting. The proposed model can learn from the price history of different assets, and it seems to produce fairly accurate forecasts.
\end{abstract}

\section{Introduction}
Convolutional neural networks have shown great results in time series forecasting. However, the applications so far, as time series forecasting in general, focused mainly on predicting the mean. This article presents a convolutional neural network for forecasting quantiles.\\
The QCNN model is applied to Value at Risk ($VaR$) forecasting. $VaR$ is a quantile of the loss distribution, thus it may be forecasted using quantile regression---either simple or deep quantile regression.\\
Deep neural networks are powerful machine learning algorithms, but they require large amounts of data. It means that the price history of a single asset may not be enough to properly train the network. Here we show that it might be beneficial to use data on multiple assets to forecast the Value at Risk of any single one.
\section{QCNN}
This section describes the proposed Quantile Convolutional Neural Network (QCNN) model.

\subsection{CNNs for Time Series Forecasting}
Convolutional neural networks use slided local receptive fields to find local features in the input data. It enables them to model certain data types particularly well, for example, images (with strong 2D structure) or time series (with strong 1D structure). Variables spatially or temporarily nearby are often correlated, so we should take advantage of the topology of the inputs \citep{lecun1995convolutional}. This is what convolutional neural networks can do very neatly.\\
CNNs are most often associated with images, yet they might be applied to one-dimensional data as well. Even the earliest convolutional architectures were applied to the time domain \citep{waibel1995phoneme}. Yet, several notable time series forecasting convolutional architectures were just proposed recently.\\
Recurrent neural networks (LSTMs and GRUs) are often considered the best or even the optimal neural network architectures for sequential data (including time series). However, \citet{bai2018empirical} compared generic recurrent and convolutional architectures on various sequence modeling tasks, and found that the latter might be a better choice.\\
Learning long temporal dependencies is a challenging task for CNNs, but dilations \citep{yu2015multi} can help. A dilated convolution means a convolution with holes, that is, the convolutional filter is enlarged by skipping a few points. The $l^{th}$ layer’s dilation rate can be set to $2^{l-1}$ in order to allow an exponential growth in the effective receptive field: by increasing the number of layers, we can exponentially increase the time horizon that the network can see. The convolution should also be causal, meaning that outputs can never depend on future inputs.\\
The WaveNet model uses dilated causal convolutions for generating audio waveforms \citep{oord2016wavenet}. \citet{borovykh2017conditional} used an adaptation of the WaveNet for time series forecasting, and found it an easy to implement and time-efficient alternative to recurrent networks.\\
QCNN is a one-dimensional dilated causal convolutional network with an appropriately chosen quantile loss function.

\subsection{Quantile Regression}
While simple least squares regressions estimate the conditional mean of a given variable, quantile regressions estimate conditional quantiles \citep{koenker2001quantile}. This requires a new objective (\ref{eq:qcnn_loss}). Mean squared error targets the mean, this targets arbitrary $\theta$-quantiles. Thus, we should just change loss function.\\
\begin{equation} \label{eq:qcnn_loss}
\theta \sum_{y_i \geq \hat{q}_i}(y_i - \hat{q}_i) + (\theta-1) \sum_{y_i < \hat{q}_i}(y_i - \hat{q}_i)
\end{equation}
\subsection{Training}
The loss function can be minimized using any proven optimizer algorithm (e.g., stochastic gradient descent or its variants).\\
Neural networks provide greater flexibility than simple linear models. However, a single time series might not be enough to exploit this flexibility. Thus, we may expect QCNN to work better when trained jointly on a set of similar time series.

\section{Empirical Study}
This section presents the application of QCNN to Value at Risk forecasting.
\subsection{Value at Risk Forecasting}
Value at Risk ($VaR$) is an important risk measure in finance. It aims to find a realistically worst-case outcome in the sense that anything worse happens with a given (small) probability \citep{shin2010risk}. $VaR$ is the worst loss over a horizon that will not be exceeded with a given level of confidence \citep{jorion2000value}. It is reported as a positive number, by convention.\\
Mathematically, the $VaR$ of $X$ with a confidence level $1-\theta$ can be defined as the quantile function of $-X$ at $1-\theta$ (\ref{eq:var}).
\begin{equation} \label{eq:var}
VaR_{\theta}(X) = F_{-X}^{-1}(1-\theta) = -\inf\left \{ x \in \mathbb{R}: \theta \leq F_X(x) \right \}
\end{equation}
$VaR$ can be estimated in many ways. A usual procedure estimates the return variance, and assumes a normal (or t) distribution to compute the quantile estimates. ARCH/GARCH models are often used to analyze and forecast volatility \citep{engle2001garch}. However, these distributional assumptions are often invalid. Efforts have been made to relax them, see, for example, \citet{hull1998value} or \citet{glasserman2002portfolio}. One of the great advantages of the quantile regression approach is that it does not require any such assumptions.\\
Quantile regression methods are often applied to Value at Risk forecasting. \citet{engle1999caviar} proposed the CAViaR (Conditional Value at Risk By Quantile Regression) model. \citet{taylor2000quantile} applied a quantile regression neural network approach to $VaR$ estimation. \citet{chen2002application} found that the quantile regression approach outperforms the variance-covariance approach. \citet{taylor2007using} developed an exponentially weighted quantile regression. \citet{white2015var} proposed a vector autoregressive extension to quantile models. \citet{xu2016quantile} applied a quantile autoregression neural network (QARNN) as a flexible generalization of quantile autoregression \citep{koenker2006quantile}. \citet{yan2018parsimonious} used a long short-term memory neural network in a quantile regression framework to learn the tail behavior of financial asset returns. Just to mention a few notable studies.\\
$VaR$ deals with rare events, and rare events produce small data. The higher confidence level we choose, the fewer loss events we have to learn from. Data volume is thus a crucial issue in Value at Risk forecasting. It is, therefore, beneficial to use several stocks’ data to learn to forecast $VaR$. More stocks have experienced more extreme events, and we may expect them to have similar sources. If they do so, then a joint $VaR$-forecasting model might outperform individual models.
\subsection{Data}
Our stock market dataset was obtained from Kaggle\footnote{https://www.kaggle.com/qks1lver/amex-nyse-nasdaq-stock-histories}. We have randomly chosen 100 stocks listed on NASDAQ, NYSE, and AMEX in the 10-year period under study (2009-01-01 to 2018-12-31). Daily logarithmic returns were computed and fed to the algorithms. The last 30\% of each time series was used as a test set, that is, about the last 3 years.
\subsection{Models and Experiments}
VaRs are forecasted for 3 confidence levels (95\%, 99\%, and 99.9\%), using 5 different methods:
\begin{itemize}
  \item a constant quantile estimate,
  \item a GARCH(1,1) with normal distribution,
  \item a linear quantile regression,
  \item a QCNN,
  \item a joint QCNN trained on all available training data.
\end{itemize}
The QCNN network contains 6 causal convolutional layers (each with 8 filters of kernel size 2, with $relu$ activation, and exponentially increasing dilation rates), and a convolutional layer with a single filter of kernel size 1 and a linear activation.\\
The inputs are overlapping 128-step sequences extracted from the time series of stock returns. The output sequences are the inputs shifted by 1 step, so that at any time we predict $VaR$ one day ahead.\\
The dataset was scaled by subtracting the mean and dividing by the standard deviation, and it was fed to the algorithms in 128-batches. The QCNN models were trained for 128 epochs, using the adadelta \citep{zeiler2012adadelta} optimizer.\\
The linear model is an autoregression using 4 lags of the target variable.\\
The constant quantile estimate was computed using linear interpolation (\ref{eq:quantile}).
\begin{equation} \label{eq:quantile}
x_{[i]} + (i - [i])(x_{[i]+1} - x_{[i]}),\quad i = (N-1)\theta + 1
\end{equation}
\subsection{Evaluation}
We apply the Dynamic Quantile test of \citet{engle1999caviar}. A $Hit$ variable (\ref{eq:dq_hit}) is constructed, which takes the value $1-\theta$ at $VaR$ exceedances, and takes $-\theta$ else, and so has an expected value of zero. This variable is regressed on $X$ (\ref{eq:dq_regression}), which may contain $Hit$'s past lags, the Value at Risk (and its lags), and possibly further variables. This regression should have no explanatory power, so we test the hypothesis $H_0: \delta = 0$. Applying the central limit theorem, we can construct an asymptotically chi-square distributed test statistic (\ref{eq:dq_teststat}). $q$ denotes the number of input variables, $t$ denotes the time steps.
\begin{equation} \label{eq:dq_hit}
Hit_t = I(y_t < -VaR_t) - \theta
\end{equation}
\begin{equation} \label{eq:dq_regression}
Hit = X\delta + u
\end{equation}
\begin{equation} \label{eq:dq_teststat}
\frac{Hit'X[X'X]^{-1}X'Hit}{\theta(1-\theta)} \overset{a}{\sim} \chi^2(q)
\end{equation}
This test is simple and easy to implement. Here we use $VaR$, and 3 lags of $Hit$ as input variables.\\
The average estimated $VaR$ values and $VaR$ exceedance rates are also reported for a more detailed assessment of model performance.

\subsection{Results}
The means, medians, and standard deviations of $VaR$ exceedances, the average rejection rates of the DQ test at two different significance levels, and the average $VaR$ values are reported in Tables \ref{table:var_005}, \ref{table:var_001}, and \ref{table:var_0001}. Zero-exceedance forecasts are assigned a 0 p-value for the DQ test. The DQ test results are not reported for the 99.9\% $VaR$ level.

\begin{table}[h!]
\centering
\begin{tabular}{ |c|c|c|c|c|c|c| }
 \hline
 \multirow{2}{*}{} & \multicolumn{3}{c|}{Exceedances} & \multicolumn{2}{c|}{DQ Test Rejections} & VaR \\
 \cline{2-7}
 & Mean & Median & SD & 0.01 & 0.05 & Mean \\
 \hline
 Constant&0.0399&0.0358&0.0291&0.5300&0.6600&0.0727 \\
 GARCH&0.0347&0.0338&0.0134&0.2100&0.4100&0.0739 \\
 QR&0.0409&0.0344&0.0293&0.6200&0.7200&0.0689 \\
 QCNN&0.1269&0.1245&0.0362&0.7400&0.8000&0.0423 \\
 Joint QCNN&0.0433&0.0444&0.0101&0.0500&0.1600&0.0583 \\
 \hline
\end{tabular}
\caption{95\% VaR forecasts}
\label{table:var_005}
\end{table}

\begin{table}[h!]
\centering
\begin{tabular}{ |c|c|c|c|c|c|c| }
 \hline
 \multirow{2}{*}{} & \multicolumn{3}{c|}{Exceedances} & \multicolumn{2}{c|}{DQ Test Rejections} & VaR \\
 \cline{2-7}
 & Mean & Median & SD & 0.01 & 0.05 & Mean \\
 \hline
 Constant&0.0084&0.0061&0.0117&0.3800&0.4100&0.1935 \\
 GARCH&0.0133&0.0119&0.0066&0.1400&0.2000&0.1039 \\
 QR&0.0097&0.0066&0.0125&0.5400&0.5900&0.1852 \\
 QCNN&0.0599&0.0576&0.0280&0.9500&0.9600&0.0676 \\
 Joint QCNN&0.0115&0.0119&0.0056&0.0900&0.1600&0.1023 \\
 \hline
\end{tabular}
\caption{99\% VaR forecasts}
\label{table:var_001}
\end{table}

\begin{table}[h!]
\centering
\begin{tabular}{ |c|c|c|c|c| }
 \hline
 \multirow{2}{*}{} & \multicolumn{3}{c|}{Exceedances} & VaR \\
 \cline{2-5}
 & Mean & Median & SD & Mean \\
 \hline
 Constant&0.0023&0.0007&0.0052&0.3764 \\
 GARCH&0.0060&0.0053&0.0041&0.1375 \\
 QR&0.0041&0.0013&0.0075&0.3630 \\
 QCNN&0.0249&0.0212&0.0186&0.1079 \\
 Joint QCNN&0.0023&0.0013&0.0026&0.2062 \\
 \hline
\end{tabular}
\caption{99.9\% VaR forecasts}
\label{table:var_0001}
\end{table}

The single-stock QCNN overshot the targeted $VaR$-exceedance rates, in some cases by orders of magnitude. All other methods produced more conservative, and seemingly more reliable forecasts. Even the constant $VaR$ estimate produced mean exceedance rates close to the desired levels. However, the joint QCNN achieved similar exceedance accuracy with lower average $VaR$ estimates. It means that while a simple historical quantile estimate might be enough to set a budget against a realistically worst-case outcome, the QCNN gives a cheaper solution. Also, the joint QCNN produced $VaR$ exceedance rates with consistently lower standard deviation than the benchmark methods. The Dynamic Quantile test is rejected for much fewer stocks in case of the joint QCNN than any other method, which also justifies that this one produces the highest quality $VaR$ estimates.\\
The experiments were repeated for the previous 10 years’ data (1999-2008) with a different set of randomly chosen stocks, and the results were quite similar.

\section{Summary}
This article proposed a one-dimensional convolutional neural network architecture for Value at Risk forecasting. The model takes as input a series of observations (returns), and directly makes one-step ahead forecasts for arbitrary quantiles. When trained jointly on multiple time series (that is, using data on multiple assets), the proposed method seems to perform well.

\bibliography{bib}

\end{document}